%% file: loner.tex
\documentclass[letterpaper, 10 pt, journal, twoside]{IEEEtran}
\IEEEoverridecommandlockouts

\usepackage{times}
\usepackage{makecell}

\usepackage{etoolbox}
\makeatletter
\patchcmd{\@makecaption}
  {\scshape}
  {}
  {}
  {}
\makeatother

\usepackage{tikz}
\newcommand*\emptycirc[1][.6ex]{\tikz\draw (0,0) circle (#1);} 
\newcommand*\halfcirc[1][.6ex]{%
  \begin{tikzpicture}
  \draw[fill] (0,0)-- (90:#1) arc (90:270:#1) -- cycle ;
  \draw (0,0) circle (#1);
  \end{tikzpicture}}
\newcommand*\fullcirc[1][.6ex]{\tikz\fill (0,0) circle (#1);} 

\usepackage[numbers]{natbib}
\usepackage{multicol}
\usepackage{multirow}
\usepackage{graphicx}
\usepackage[bookmarks=true]{hyperref}
\usepackage{amsmath}
\usepackage{amssymb}
\usepackage{float}
\usepackage{pifont}
\usepackage{placeins}
\usepackage{mathdots}
\usepackage[caption=false,font=normalsize,labelfont=sf,textfont=sf]{subfig}
\usepackage{xspace}
\usepackage{svg}
\usepackage{rotating}
\usepackage{fontawesome}
\usepackage{pifont}
\newcommand{\fail}{\xmark\xspace}

\newcommand{\R}{\mathbb{R}}
\newcommand{\algname}{LONER\xspace}
\newcommand{\na}{-}

\newcommand{\verttab}[1]{\begin{tabular}{@{}c@{}}\rotatebox[origin=c]{90}{#1}\end{tabular}}

\newcommand{\xmark}{\ding{55}}%

\newcommand{\x}{\mathbf{x}}

\begin{document}

\title{\algname: \textbf{L}iDAR \textbf{O}nly \textbf{Ne}ural \textbf{R}epresentations for Real-Time SLAM}

\author{Seth Isaacson$^{*,1}$,
Pou-Chun Kung$^{*,1}$,
Mani Ramanagopal$^1$,
Ram Vasudevan$^1$, and Katherine A. Skinner$^1$

\thanks{Manuscript received: May 30, 2023; Revised August 28, 2023; Accepted October 1, 2023.}%Use only for final RAL version
\thanks{This paper was recommended for publication by Editor Javier Civera upon evaluation of the Associate Editor and Reviewers' comments.}
\thanks{This work was supported by the Ford Motor Company via the Ford-UM Alliance under award N028603.}
\thanks{$^1$S. Isaacson, P. Kung, M. Ramanagopal, R. Vasudevan, and K. A. Skinner are with the Department of Robotics, University of Michigan, Ann Arbor, MI 48109. \texttt{\{sethgi, pckung, srmani, ramv, kskin\}@umich.edu}.}
\thanks{Digital Object Identifier (DOI): see top of this page.}
\thanks{$^*$These two authors contributed equally to this work.}
}

\markboth{IEEE Robotics and Automation Letters. Preprint Version. Accepted October, 2023}{Isaacson \MakeLowercase{\textit{et al.}}: LONER} 

%This paper was produced by the IEEE Publication Technology Group. They are in Piscataway, NJ.}% <-this % stops a space



\maketitle

\begin{abstract}
This paper proposes \textit{\algname}, the first real-time LiDAR SLAM algorithm that uses a neural implicit scene representation. 
Existing implicit mapping methods for LiDAR show promising results in large-scale reconstruction, but either require groundtruth poses or run slower than real-time.
In contrast, \algname uses LiDAR data to train an MLP to estimate a dense map in real-time, while simultaneously estimating the trajectory of the sensor.\
To achieve real-time performance, this paper proposes a novel information-theoretic loss function that accounts for the fact that different regions of the map may be learned to varying degrees throughout online training. % different regions of the map having varying degrees of uncertainty during online operation.
The proposed method is evaluated qualitatively and quantitatively on two open-source datasets. 
This evaluation illustrates that the proposed loss function converges faster and leads to more accurate geometry reconstruction than other loss functions used in depth-supervised neural implicit frameworks. 
% In addition, this paper shows that the proposed trajectory estimation method is competitive with state-of-the-art LiDAR SLAM methods.
Finally, this paper shows that \algname estimates trajectories competitively with state-of-the-art LiDAR SLAM methods, while also producing dense maps competitive with existing real-time implicit mapping methods that use groundtruth poses. %Project page: \href{https://umautobots.github.io/loner}{https://umautobots.github.io/loner}

% Unlike existing LiDAR SLAM methods, \algname estimates dense maps that support novel view synthesis.
% Finally, \algname produces maps that are competitive with a state-of-the-art real-time implicit LiDAR mapping method using groundtruth poses. %compare with SHINE mapping
%Finally, \algname produces maps with up to 10\% greater accuracy than the state-of-the-art real-time implicit mapping method. %compare with SHINE mapping
%Hence, we show \algname to be the first LiDAR SLAM method that offers real-time trajectory estimation and a dense neural implicit scene representation.

%An open-source implementation is released.

\end{abstract}


\section{Introduction} \label{introduction}

\IEEEPARstart{N}{eural} implicit scene representations, such as Neural Radiance Fields (NeRFs), offer a promising new way to represent maps for robotics applications \cite{nerf}.
Traditional NeRFs employ a Multi-Layer Perceptron (MLP) to estimate the radiance and volume density of each point in space, enabling dense scene reconstruction and novel view synthesis.
The learned scene representation has several advantages over conventional map representations, such as point clouds and occupancy grids.
First, because the domain of the NeRF is continuous and does not enforce discretization, any point in the scene can be queried for occupancy.
The continuity of the scene can be exploited to solve a variety of robotics problems.
For example, as demonstrated in \cite{nerf-navigation}, a motion planner can integrate the volume density along a proposed trajectory to evaluate the likelihood of a collision.
Other benefits include the ability to produce realistic renders of the scene \cite{nerf}. 
Further, NeRFs can be used to estimate uncertainty of renders to enable view selection for active exploration \cite{active-nerf}.
%Other benefits include the ability to produce realistic renders and to create a mesh representation of the implicit geometry.
% a SLAM algorithm that leverages a neural scene representation has advantages for robotics applications.
This paper advances neural implicit scene representations for robotics applications. Specifically,  we introduce the first real-time LiDAR-only SLAM algorithm that achieves accurate pose estimation and map reconstruction and learns a neural implicit representation of a scene.

\begin{figure}[t!]
    \centering
\includegraphics[width=0.94\linewidth]{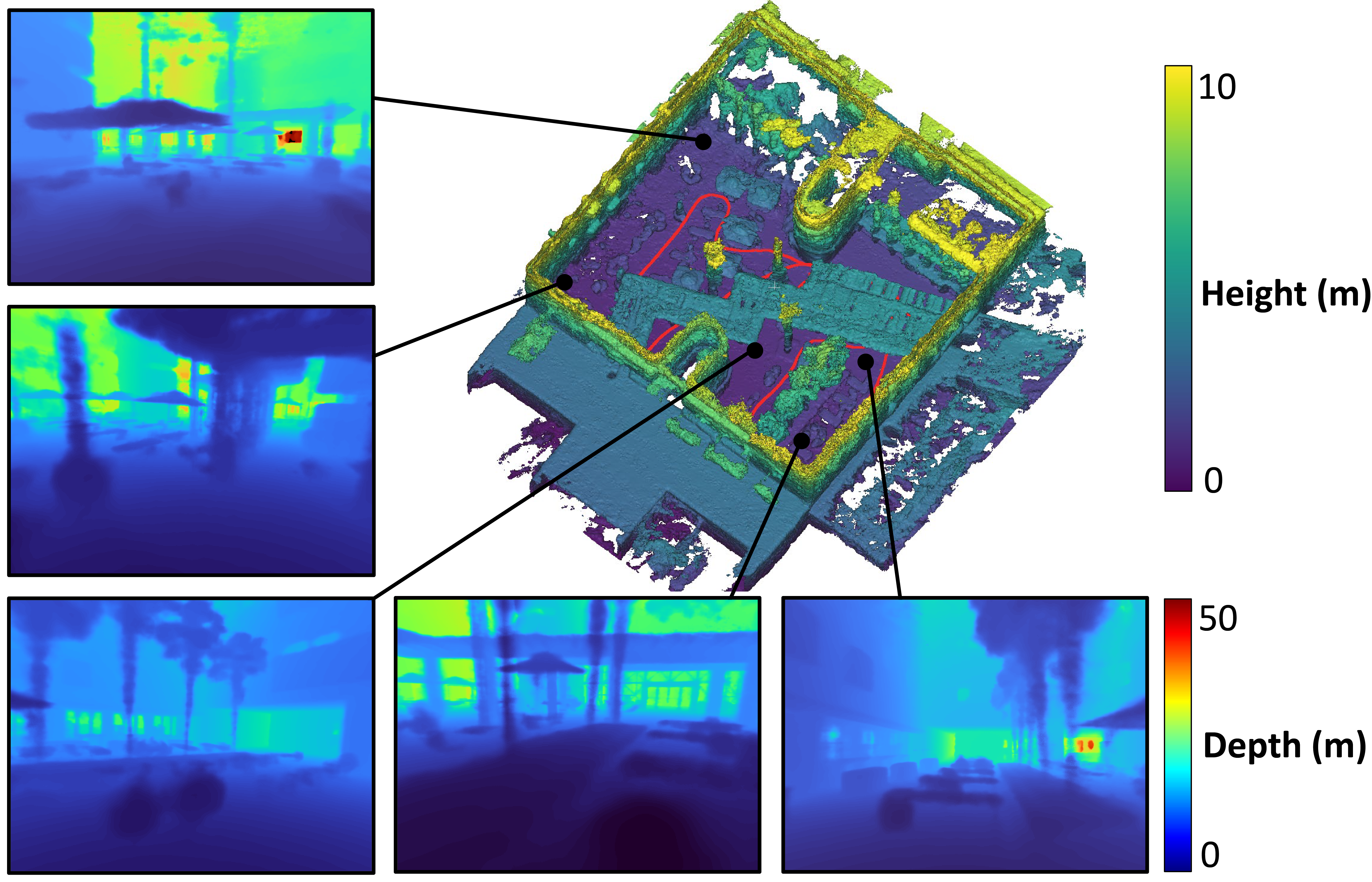}
    \caption{LONER reconstruction on a courtyard scene \cite{fusion-portable}. The top-right is a mesh reconstruction with the estimated trajectory in red. The surrounding images are rendered depth images from novel views outside of the training trajectory, demonstrating \algname's ability to reconstruct dense novel views of an environment.}
    \label{fig:fig1}
\end{figure}

Several recent papers have proposed real-time NeRF-based visual SLAM systems using monocular or RGB-D cameras~\cite{imap, niceslam, nerf-slam}. These systems demonstrate impressive performance on indoor scenes. %Because these systems use monocular or RGB-D cameras, they are unsuitable for outdoor and large-scale scenarios due to inherent limitations in the sensors. 
For outdoor environments, prior work has focused on using neural implicit representations for LiDAR to enable dense 3D reconstruction and novel view synthesis for large-scale scenes \cite{urf, cloner, shinemapping}. 
Recent methods have even shown promising results for LiDAR localization and mapping with neural implicit frameworks in large-scale outdoor scenes \cite{nerfloam, yan2023}. 
Still, these LiDAR-supervised algorithms do not operate in real-time, which is necessary for robotics applications.

% This paper proposes \algname, the first real-time neural implicit LiDAR SLAM method. 
% Unlike existing neural implicit algorithms, \algname can operate in outdoor environments and provides accurate online state estimation and dense mapping capabilities. 
% To achieve this, this paper introduces a novel loss function that we show leads to faster convergence and more accurate reconstruction than existing loss functions. 
% Finally, we perform extensive experiments on two open-source datasets to demonstrate that our proposed method runs in real-time and outperforms baselines in terms of dense mapping and trajectory estimation. 
% Figure~\ref{fig:fig1} illustrates sample reconstruction results on the Fusion Portable dataset \cite{fusion-portable}.

\noindent The contributions of this paper are as follows: 
\begin{enumerate}
    \item We propose the first real-time neural implicit LiDAR SLAM method, which adapts to outdoor environments and provides accurate online state estimation. 
    \item We introduce a novel loss function that leads to faster convergence and more accurate reconstruction than existing loss functions. 
\end{enumerate}
We demonstrate that our proposed method, \algname, runs in real-time and estimates both trajectories and maps more accurately than baselines. Figure~\ref{fig:fig1} shows the reconstruction results on the Fusion Portable dataset \cite{fusion-portable}. A project page is available at \href{https://umautobots.github.io/loner}{https://umautobots.github.io/loner}.

The remainder of this paper is organized as follows: In Section \ref{relatedworks}, we review related work. In Section \ref{method}, we describe \algname. In Section \ref{experiments}, we evaluate \algname, and in Section \ref{conclusion}, we conclude and discuss both limitations and future work.

% \IEEEpubidadjcol

\section{Related Works} \label{relatedworks}
%This section reviews traditional LiDAR SLAM algorithms, introduces NeRF-based SLAM methods and works that incorporate LiDAR to train neural scene representations, summarizes the loss functions used in existing depth-supervised NeRFs, and discusses why these losses are not suitable for SLAM.

\subsection{LiDAR SLAM} \label{related-slam}
% Seth

% Visual SLAM has been an extremely active research area over the past two decades. The first widely successful approaches leveraged hand-designed features which are tracked between frames \cite{ptam, orbslam}. These features are then triangulated and their locations jointly optimized with the camera poses in a bundle adjustment optimization. More recent approaches have built on these early successes by incorporating a variety of sensing modalities, multi-session SLAM, and other features to improve performance \cite{orbslam2, orbslam3}. These approaches achieve excellent trajectory accuracy, but produce only sparse maps which are typically not useful for reconstruction tasks.

% Dense Visual SLAM algorithms do not rely solely on features, and are able to produce dense maps. Early approaches such as DTAM \cite{dtam} have lead to algorithms such as DSO \cite{dso} and LSD-SLAM \cite{lsd-slam}, all of which operate by optimizing a photometric loss. Other approaches such as SVO \cite{svo} hybridize features and direct methods. 

% Recently, deep learning has attracted more attention in the Visual SLAM literature. Learned features have been shown to significantly outperform their hand-designed counterparts in feature-based methods \cite{lift, loftr, rwt-slam, lf-net}. Further, DROID-SLAM solves both the SLAM front-end and back-end as machine learning problems \cite{droid-slam}.
LiDAR SLAM has been an active research area over the past several decades \cite{loam, suma, lego-loam, lio-mapping, lio-sam}. The primary goal of these methods is to estimate the trajectory of the ego vehicle. Modern methods such as LeGO-LOAM estimate motion by aligning features extracted from consecutive scans, then accumulate LiDAR scans to build a map \cite{loam, lego-loam}. 
These works primarily focus on accurate trajectory estimation, and thus creating dense, realistic maps is not a focus of the approach. In contrast, our method aims to achieve similar or better trajectory estimation while also estimating dense maps.
% A common approach is to align consecutive LiDAR scans to estimate the relative motion of the sensor. %Modern methods such as LiDAR Odometry and Mapping (LOAM) first extract features from LiDAR point clouds by evaluating the curvature of the cloud \cite{loam}. Then, associated features are aligned to estimate relative motion. Maps are then built by accumulating LiDAR scans.
% LiDAR Odometry and Mapping (LOAM) \cite{loam} first extracts features from LiDAR point clouds. Associated features are then aligned to estimate relative motion \cite{loam}. Pointcloud-based maps are then built by accumulating LiDAR scans.LeGO-LOAM builds upon LOAM by segmenting input point clouds to improve feature matching. This results in state-of-the-art trajectory accuracy in large-scale outdoor environments. \cite{lio-mapping, lio-sam} further introduce inertial sensors to improve SLAM performance, but using external sensors is out of the scope of this paper. While previous LiDAR SLAMs only provide a point cloud-based map, our proposed method constructs a dense neural implicit scene reconstruction.
%Our proposed method poses trajectory estimation as an online optimization problem using a neural implicit map representation.

\subsection{Real-time NeRF-based SLAM} \label{related-implicit}

NeRFs use images of a scene captured from known camera poses to train an MLP to predict what the scene will look like from novel poses~\cite{nerf}.
While originally developed for offline use with known camera poses, NeRFs have recently been used to learn an implicit scene representation in RGB and RGB-D SLAM frameworks~\cite{imap, niceslam, nerf-slam}. 
By representing the scene with a NeRF, these algorithms perform both tracking and mapping via gradient descent on an MLP or related feature vectors. 
For example, iMAP represents the scene as a single MLP~\cite{imap}. 
Each RGB-D frame is first tracked by fixing the MLP weights and optimizing the camera poses. 
Then, the new information is incorporated into the map by jointly optimizing the MLP and pose estimates.
iMAP shows promising results in small scenarios but does not scale to larger or outdoor scenes.
NICE-SLAM replaces iMAP's simple MLP with a hierarchical feature grid combined with MLP decoders \cite{niceslam}. 
This approach demonstrates better scalability than the single MLP used in iMAP, but NICE-SLAM still only works in indoor scenarios. 
Additionally, NeRF-SLAM uses DROID-SLAM \cite{droid-slam} as the tracking front-end, which allows them to use a probabilistic volumetric NeRF to perform uncertainty-aware mapping and pose refinement \cite{nerf-slam}. 
Recently, several more papers have introduced architectures and encodings to improve neural-implicit SLAM's memory efficiency, computation speed, and accuracy \cite{meslam, co-slam, eslam, vox-fusion}.
Our method extends these recent advances to leverage implicit scene representation for real-time LiDAR-only SLAM, which allows operation in large, outdoor environments.

\subsection{Neural Implicit Representations for LiDAR} \label{related-lidar}
While neural implicit representations were initially developed for visual applications, several works have introduced neural implicit representations for LiDAR to improve outdoor 3D reconstruction performance \cite{urf, dsnerf, cloner}. 
Urban Radiance Fields (URF) is an early example of LiDAR-integraded NeRF \cite{urf}. 
URF uses a novel Line-of-Sight (LOS) loss to improve LiDAR supervision. %, as is explained in the following section.
%The LOS loss uses a normal distribution with a decaying variance to represent LiDAR depth measurements during training. However, with a fixed variance decay rate, the convergence speed for all rays is limited by the handcrafted decay rate. In contrast, we propose a dynamic variance strategy based on JS-Divergence, which applies different variances for each ray. By doing this, we demonstrate improved efficiency when integrating LiDAR supervision into NeRF.
CLONeR uses LiDAR and camera data to train two decoupled MLPs, one of which learns scene structure and the other of which learns scene color \cite{cloner}. 
CLONeR combines the decoupled NeRF with occupancy-grid enabled sampling heuristics and URF's Line-of-Sight loss to enable training with as few as two input views \cite{cloner}. 
Both URF and CLONeR require known sensor poses and assume offline training.
In contrast, our proposed method performs real-time LiDAR SLAM that both reconstructs 3D environments and estimates sensor poses for sequential input data.

%shine mapping

In \cite{yan2023}, a method is introduced that inputs LiDAR scans and approximate poses, then uses a novel occlusion-aware loss function to jointly optimize the poses and a NeRF. 
This work assumes a-priori availability of all data. 
Thus, it can be effectively viewed as a LiDAR-based structure-from-motion algorithm, whereas we present a full SLAM algorithm.
Recently, SHINE Mapping presented a LiDAR mapping method based on neural signed distance function (SDF) and sparse feature embedding~\cite{shinemapping}. 
While this embedding helps scale to large scenes, in a real-time configuration, it presents a trade-off between hole-filling and overall map resolution. Our method instead uses a dense feature embedding, which enables improved performance across both hole-filling capability and map resolution.
NeRF-LOAM extends this to a LiDAR SLAM system and proposes a dynamic voxel embedding generation strategy to adapt to large-scale scenarios \cite{nerfloam}. However, it does not operate in real-time. %However, it is an offline SLAM method without real-time capability.
%However, we found the choice in voxel resolution and number of training samples is a trade-off for those voxel-based approaches in real-time configuration. The map is either sparse or with low resolution within a limited training time.

\begin{figure*}[h!t!]
    \centering
    \includegraphics[width=0.9\textwidth]{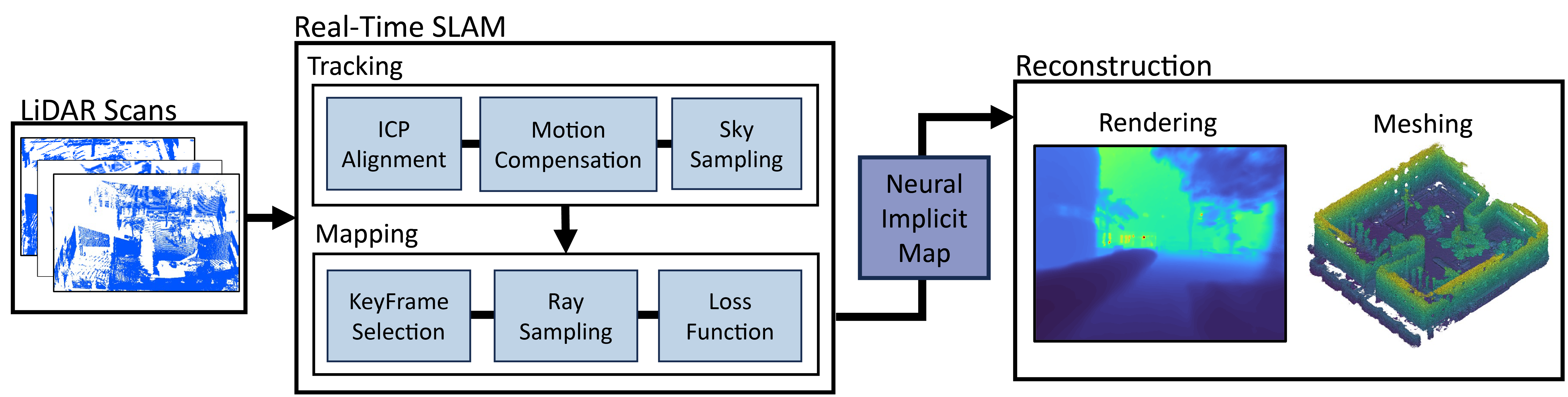}
    \caption{\algname system overview. Incoming scans are decimated then tracked with ICP, after which the sky is segmented from the scene geometry. Selected scans are chosen as KeyFrames. Each map update includes the current KeyFrame and randomly selected past KeyFrames. Our novel loss function is used to update the poses and MLP weights. The resulting implicit map can be rendered offline to a variety of formats including depth images and meshes.}
    \label{fig:method-architecture}
\end{figure*}

\subsection{Loss for Depth-supervised NeRF} \label{related-loss}

Depth-supervised NeRF frameworks, such as those that use RGB-D sensors, typically use the difference between rendered and sensed depth as a loss to learn geometry from 2D images by volumetric rendering \cite{imap, niceslam}. Other works use depth measurements directly in 3D space to perform depth-supervision~\cite{dsnerf, urf, cloner, yan2023}. 
The Binary Cross-Entropy (BCE) loss proposed in \cite{yan2023} reasons about occluded objects, but does not consider measurement uncertainty. 
%The KL divergence loss presented by DS-NeRF\cite{dsnerf} and Line-Of-Sight (LOS) loss introduced by \cite{urf} represent depth measurements as normal distributions to create continuity and account for depth uncertainty. 
The KL divergence loss presented by DS-NeRF \cite{dsnerf} and Line-Of-Sight (LOS) loss introduced by URF \cite{urf} approximate each LiDAR ray's termination depth as a normal distribution centered at the measured depth.
The variance of the distribution is correlated with a margin parameter $\epsilon$. 
The loss functions encourage the network to predict weights along a ray equal to the PDF of the normal distribution. 
%Unlike the KL loss \cite{dsnerf} that uses a fixed margin throughout training, the LOS loss \cite{urf} shows that exponentially decaying the margin $\epsilon$ leads to better reconstruction accuracy than using a fixed margin.
While the KL loss leaves the variance fixed during training, \cite{urf} shows that decaying $\epsilon$ during training improves reconstruction accuracy when using the LOS loss.

While uniformly decaying a margin is successful offline, using a single margin for all rays is unsuitable for real-time SLAM, which has incremental input and limited training samples. 
Using a uniform margin can force the NeRF model to forget learned geometry when adding a new LiDAR scan and can cause slower convergence.
%Also, we found the choice of the margin decay rate is a trade-off between hole-filling capability and reconstruction accuracy when having limited training time, as shown in Figure~\ref{fig:depth_image_loss_comparison}. 
% This paper proposes a novel JS divergence-based dynamic margin strategy, which implicitly balances this trade-off.
Therefore, this paper proposes a novel dynamic margin loss that applies a different margin for each ray. 
We demonstrate the proposed loss function leads to better 3D reconstruction than previous loss functions within fewer training samples, and enables real-time performance.

\section{Method} \label{method}

This section provides a high-level overview of our proposed system, \algname, before explaining each component in detail.

% \begin{figure}
%     \centering
%     \includegraphics[width=0.9\columnwidth]{figures/Architecture.png}
%     \caption{\algname comprises a tracking thread that estimates poses in real-time and a lower-frequency mapping thread that updates the NeRF map.}
%     \label{fig:method-architecture}
% \end{figure}

\subsection{System Overview}\label{system-overview}
An overview of \algname is shown in Fig.~\ref{fig:method-architecture}. As is common in the SLAM literature \cite{imap, niceslam, orbslam}, the system comprises parallel threads for tracking and mapping. The tracking thread processes incoming scans and estimates odometry using ICP. \algname is designed for use without an IMU, so ICP uses the identity transformation as an initial guess. In parallel and at a lower rate, the mapping thread uses the current scan and selected prior scans as KeyFrames, which are used to update the training of the neural scene representation.

\subsection{Tracking} \label{method-tracking}
Incoming LiDAR scans are decimated to a fixed frequency of 5Hz. The relative transform $P_{i-1, i} \in SE(3)$ from the previous scan to the current scan is estimated using Point-to-Plane ICP \cite{icp}. We use ICP instead of inverse-Nerf since it offers good real-time performance while allowing GPU resources to be dedicated to the mapping module. The ICP estimate is later refined in our mapping optimization. The LiDAR pose $\x_i \in SE(3)$ is then estimated as $\hat \x_i = \hat \x_{i-1} \cdot P_{i-1, i}$. Given the previous and current pose, LiDAR scans are motion-compensated by assuming constant velocity motion between scans.

\subsection{Implicit Map Representation} \label{method-scene}
The scene is represented as an MLP with the hierarchical feature grid encoding from \cite{instant-ngp}. During online training, the parameters $\Theta$ of the MLP and the feature grid are updated to predict the volume density $\sigma$ of each point in space. To train the network and estimate depths, we follow the standard volumetric rendering procedure \cite{nerf}. In particular, for a LiDAR ray $\vec r$ with origin $\vec o$ and direction $\vec d$, we choose distances $t_i \in [t_{near}, t_{far}]$ to create $N_S$ samples $s_i = \vec o + t_i \vec d$. LiDAR intrinsics dictate $t_{near}$, while $t_{far}$ depends on the scale of the scene. The feature grid and MLP, collectively $\mathcal{F}(s_i; \Theta)$, are queried to predict the occupancy state $\sigma_i$. Then, weights transmittances $T_i$ and weights $w_i$ are computed according to:

\begin{align}\label{eq:rendering}
% \sigma_i &= \mathcal{F}(s_i; \Theta) \\
T_i &= \exp(- \sum_{j=1}^{i-1} \sigma_j \delta_j) \\
w_i &= T_i(1 - \exp(-\sigma_i\delta_i)) \label{eq:weight_def}
\end{align}

\noindent where $\delta_j = t_{j+1} - t_j$, and $\sigma_i$ is the density at sample $s_i$ predicted by the MLP. The weights $w_i$ are used by the loss function and represent the probability that the ray terminates at each point. Therefore, the expected termination depth of a ray $\hat{D}(\vec r)$ can be estimated as
\begin{align} \label{eq:depth-pred}
\begin{split}
    \hat{D}(\vec r) &= \sum_{i=1}^N w_i t_i.
\end{split}
\end{align}

\subsection{Mapping} \label{method-mapping}
The mapping thread receives LiDAR scans from the tracking thread and determines whether to form a KeyFrame. If the scan is accepted, the map is jointly optimized with the poses.

\subsubsection{KeyFrames} \label{method-keyframes}
KeyFrames are selected temporally: if $t_{KF}$ has passed since the previous KeyFrame, a new KeyFrame is added. Each time a KeyFrame is accepted, the optimizer is updated. $N_W$ total KeyFrames are used in the update, including the current KeyFrame and $N_W-1$ random selected past KeyFrames.
%half of which are the most recent accepted KeyFrames, and the remainder of which are randomly selected from all past KeyFrames.

\subsubsection{Optimization}

Once the window of KeyFrames has been selected, the map is jointly optimized with the poses of KeyFrames in the optimization window. For a KeyFrame $\mathbf{KF}_i$ with estimated pose $\hat \x_i$ in the world frame, a twist vector $\hat \xi_i \in \R^6$ is formed to be used as the optimization variable. Specifically, $\hat \xi_i = (\hat \omega_i, \hat v_i)$ where $\hat \omega_i$ is the axis-angle representation of the rotation component of $\hat \x_i$, and $\hat v_i$ is the translation component. In the forward pass, this vector is converted back into a pose $\hat \x_i$ and used to compute the origin of rays. $N_R$ rays are sampled at random from the LiDAR scan, and $N_S$ depth samples are taken from each ray using the occupancy grid heuristic introduced by \cite{cloner}. 

In the backward pass, gradients are computed for MLP and feature grid parameters $\Theta$ and twist vectors $\hat{\xi}_i$. At the end of the optimization, the optimized twist vectors $\xi_i^*$ are converted into $SE(3)$ transformation matrices $\x_i^*$. The tracking thread is informed of this change, such that future tracking is performed relative to the optimized poses.

\begin{figure}[t!]
    \centering
    \includegraphics[width=0.95\linewidth]{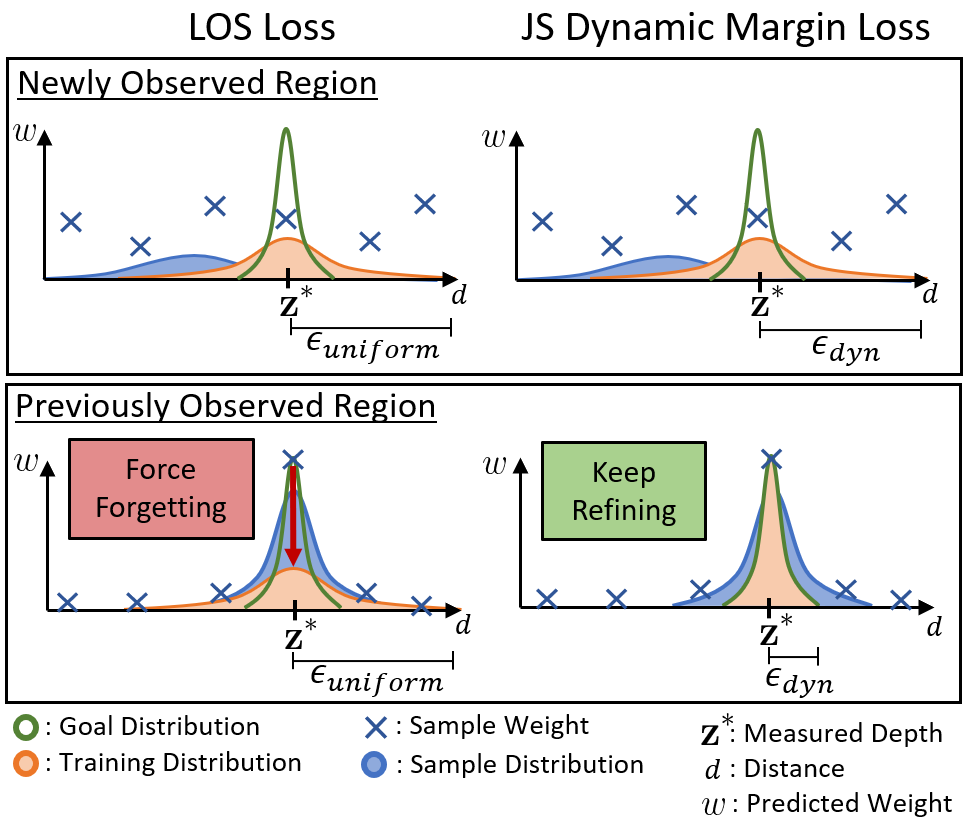}
    \caption{Illustration of the difference between the JS loss and the LOS loss. 
    The LOS loss sets a uniform margin $\epsilon$ for rays pointing to both learned and unobserved regions.
    This strategy corrupts the learned information by forcing learned regions to predict higher variances. In contrast, the proposed JS loss sets the dynamic margin $\epsilon$ for each ray depending on the similarity between goal distribution and predicted sample distribution. The JS loss sets higher margins for rays in unobserved regions to improve convergence, and sets lower margins for rays in learned regions to refine learned geometry.}
    \label{fig:method-js-loss}
\end{figure}

\subsection{JS Dynamic Margin Loss Function} \label{method-loss}

The primary loss function in our system is a novel dynamic margin loss. This is combined with terms for depth loss and sky loss as follows:
\begin{align}\label{eq:loss_summary}
\mathcal{L}(\Theta) &= \mathcal{L}_{JS} + \lambda_1 \mathcal{L}_{depth} + \lambda_2 \mathcal{L}_{sky}.
\end{align}
Each of these terms is explained below.

\subsubsection{JS Loss Formulation}
The LOS loss used by \cite{urf, cloner} uses a single margin for all rays; we use a similar formulation but introduce a novel strategy based on the Jensen-Shannon Divergence \cite{JS} to assign a unique margin to each ray. For a given LiDAR ray $\vec r$, the samples along the ray are $s_i = \vec o + t_i \vec d$, and $z^*$ denotes the measured depth along the ray. $t_i$ denotes the distance of individual training samples along the ray, and $w_i$ represents a corresponding weight prediction from an MLP, as defined in Equation \ref{eq:weight_def}. We define a truncated Gaussian distribution $\mathcal{K}_{\epsilon}$ that has a bounded domain parameterized by margin $\epsilon$, with $\mathcal{K}_{\epsilon} = \mathcal{N}(0,(\epsilon/3)^2)$ as the training distribution. Thus, target weights are given by $w_i^* = \mathcal{K}_{\epsilon}(t_i-z^*)$. The JS loss is defined as
\begin{equation}
\mathcal{L}_{JS}(\Theta) =  \underbrace{\lVert w_i^* - w_i \rVert_1}_{\text{Primary Loss}}  + \underbrace{\lVert 1 - \sum_i w_i \rVert_1}_{\text{Opacity Loss}},
\end{equation}
where the opacity loss (explained in more detail by \cite{cloner}) ensures weights along each ray sum to one and thus form a probability distribution. Note that while URF \cite{urf} uses an L2 loss to compute the LOS loss, we follow \cite{cloner} and use an L1 loss. The effect of this is discussed in Section \ref{experiments-performance}.

In \cite{urf, cloner}, the margin decays exponentially throughout training and, at each iteration, a single margin is shared by all of the rays. In contrast, we present a JS divergence-based dynamic margin that computes a unique margin for each ray to improve the training convergence and reconstruction accuracy.

In a SLAM application, continuous optimization, sparse sampling, and incremental input lead to different regions of the map being learned to varying degrees during online training.
As shown in Fig.~\ref{fig:method-js-loss}, using a uniform $\epsilon$ in the LOS loss causes forgetting in regions that have already been learned. The idea of the JS dynamic margin is to use a larger margin for rays pointing toward regions of the map with unknown geometry while using a smaller margin for rays pointing toward well-learned regions. This allows the system to learn new regions while preserving and refining learned geometry.
We use the JS divergence to measure the dissimilarity between the goal distribution and the sample distribution for each ray, which represents how well the map has learned along the ray. Learned regions have similar goal and sample distributions, which lead to smaller JS divergence.
We define a goal distribution $G = \mathcal{N}(z^*,\sigma^*)$, where $\sigma^* = \epsilon_{min}/3$. Further, we define the sample distribution $S = \mathcal{N}( \bar{\mu}_{w}, \bar{\sigma}_{w} )$, where $\bar{\mu}_{w}$ and $\bar{\sigma}_{w}$ denote mean and standard deviation of the predicted weights along a particular ray. The dynamic margin is then defined as 

\begin{equation}
\epsilon_{dyn} = \epsilon_{min} ( 1+\alpha \mathbf{J}^* )
\end{equation}

\begin{equation}
\mathbf{J}^* = 
% \left\{
% \begin{array}{lcl}
% 0, & JS(G||S)<JS_{min} \\
% JS_{max}, & JS(G||S)>JS_{max} \\
% JS(G||S), & otherwise
% \end{array}
% \right\\
\begin{cases}
0 & JS(G||S) < JS_{min} \\
JS_{max} & JS(G||S) > JS_{max} \\
JS(G||S) & \text{ otherwise, }
\end{cases}
\end{equation}

\noindent where $\alpha$ is a constant scaling parameter. $JS_{max}$ denotes the upper bound of the JS score, and $JS_{min}$ denotes a threshold for scaling. Once the JS score is smaller than $JS_{min}$, $\epsilon_{dyn}$ is equal to $\epsilon_{min}$.

\subsubsection{Depth Loss}\label{method-LOS}
As in \cite{urf}, we use the depth loss as an additional term in the loss function. The depth loss is the error between rendered depth and LiDAR-measured depth along each ray. The loss is defined as

\begin{equation}
    \mathcal{L}_{depth}(\Theta) = \lVert \hat{D}(\vec r) - z^* \rVert_2^2
\end{equation}

We found the depth loss contributes to blurry reconstruction with limited training time, but still provides good hole-filling, as shown in Fig.~\ref{fig:depth_image_loss_comparison}.
Hence, unlike \cite{urf} which weights depth loss and LOS loss equally, we down-weight the depth loss by setting $\lambda_1=5\times10^{-6}$. 

\subsubsection{Sky Loss}\label{loss:sky}

Similar to \cite{urf}, we add an additional loss to force weights on rays pointing at the sky to be zero.
While \cite{urf} segments the sky with camera-based semantic segmentation, we determine sky regions by observing holes in the LiDAR scans. First, each scan is converted to a depth image. This is then filtered via a single dilate and erode. Any points which remain empty reflect regions of the LiDAR scan where no return was received. If the ray corresponding to each of these points has a positive elevation angle in the global frame, it is determined to point to the sky. Thus, this heuristic works as long as the LiDAR is approximately level during initialization.  For sky rays, the opacity loss is not enforced. Then, for all sky rays, the following loss is computed:
\begin{equation}
    \mathcal{L}_{sky}(\Theta) = \lVert w \rVert_1.
\end{equation}

%Each of the components of the loss are explained below.

\subsection{Meshing}\label{method-rendering}
To form a mesh from the implicit geometry, a virtual LiDAR is placed at estimated KeyFrame poses. We compute weights along LiDAR rays, then bucket the weights into a 3D grid. When multiple weights fall within the same grid cell, the maximum value is kept. Marching cubes is then used to form a mesh from the result. This process runs offline for visualization and evaluation, and is not a part of online training.

\section{Experiments} \label{experiments}

This section evaluates the trajectory estimation and mapping accuracy of \algname against state-of-the-art baselines. We further evaluate the choice of loss function and perform ablation studies over key features.

\subsection{Implementation Details} \label{implementation-details}
Table~\ref{tab:configurations} provides parameters used for evaluation of \algname, which we found to generalize well across the tested datasets. All values were tuned experimentally to maximize performance while maintaining real-time operation. For all experiments, each method and configuration was run 5 times and we report the median result, as in \cite{orbslam}. The complete data from our evaluations is available on the project webpage.

\input{tables/configurations.tex}

\subsection{Baselines}
We evaluate against NICE-SLAM \cite{niceslam} and LeGO-LOAM \cite{lego-loam}, which represent state-of-the-art methods in neural-implicit SLAM and LiDAR SLAM respectively. 
Additionally, we evaluate our SLAM pipeline with the loss functions from CLONeR \cite{cloner} and URF \cite{urf}. We refer to these approaches as ``LONER w./ $\mathcal{L}_{\text{CLONeR}}$" and ``LONER w./ $\mathcal{L}_{\text{URF}}$" respectively. Finally, mapping performance is compared to SHINE mapping, which is run with groundtruth poses \cite{shinemapping}. Since NeRF-LOAM \cite{nerfloam} is a recent work and code is not yet available, it is excluded from this evaluation in favor of SHINE. Note that NeRF-LOAM does not operate in real-time.

\subsection{Datasets} \label{experiments-datasets}

We evaluate performance on two open source datasets, Fusion Portable \cite{fusion-portable} and Newer College \cite{newer-college}. Collectively, the chosen sequences represent a range of scales and difficulties.
From Fusion Portable, we select three scenes. The first sequence is MCR Slow 01, which is a small indoor lab scene collected on a quadruped. The others are Canteen Day and Garden Day, which are medium-scale semi-outdoor courtyard areas collected on a handheld platform. Both sequences contain few dynamic objects, as handling dynamic objects is left to future work. From Newer College, we evaluate on the Quad Easy sequence, which consists of two laps of a large outdoor college quad area. Because Newer College has monochrome fisheye cameras, it is incompatible with NICE-SLAM. Hence, NICE-SLAM is excluded from the Newer College results.

Note that the sequences used in testing do not have RGB-D sensors. Hence, we instead simulate RGB-D from stereo offline using RAFT optical flow estimation \cite{raft}, and run NICE-SLAM on the result. NICE-SLAM can fail to converge in these semi-outdoor scenarios in a real-time configuration, so we increased the number of samples and iterations to improve performance. We ran NICE-SLAM for 350 iterations per KeyFrame, used $2^{14}$ samples per ray, and selected a KeyFrame every 5 frames. This results in offline runtime performance. To bound the computational complexity, we set the middle and fine grid sizes to 0.64m and 0.32m respectively.

%\subsection{Evaluation Metrics} \label{experiments-metrics}

%Map metrics include accuracy (mean distance from each point in the estimated scan to each point in the groundtruth map) and completion (mean distance from each point in the groundtruth map to each point in the estimated scan). 

%Additionally, we compute precision and recall with a radius of 0.1m. For precision and recall, a true positive is defined as a point in the estimate with a corresponding point in the groundtruth map within the 0.1m radius. A false positive is a point in the estimated map with no corresponding point in the groundtruth within the radius, and a false negative occurs when there is a point in the groundtruth map with no point in the estimate within the radius. Due to occlusions in groundtruth point clouds, accuracy and precision cannot be evaluated on the Canteen and Garden sequences.

%To do this, we use the each sequence's LiDAR data and groundtruth trajectory to remove points from the groundtruth map which are more than 25cm away from the nearest LiDAR point detected by the sensor-under-test.

\subsection{Performance Analysis} \label{experiments-performance}

\subsubsection{Trajectory Tracking Evaluation}
\input{tables/tracking_comparison_compact}

Trajectory estimates from each algorithm are evaluated according to the procedure described in \cite{traj-evaluation}. We use an open-source package for these evaluations\footnote{\hyperlink{https://github.com/MichaelGrupp/evo}{https://github.com/MichaelGrupp/evo}}. Trajectories are aligned, then, the root-mean-squared absolute pose error is computed and referred to as $t_{APE}$.

Table~\ref{tab:tracking-comparison-compact} compares trajectory performance to state-of-the-art methods \cite{lego-loam, niceslam}. Our method offers performance competitive with or better than existing state-of-the-art LiDAR SLAM. On the evaluated scenes, we outperform LeGO-LOAM except for on Newer College Quad, which is the largest and most open sequence. Even on Quad, our estimated trajectory is within millimeters of the LeGO-LOAM result. Additionally, unlike LeGO-LOAM, our method creates dense implicit maps of the scene.
On the MCR sequence, NICE-SLAM successfully estimated a trajectory four of five times. The resulting trajectories were reasonable, but not competitive with the LiDAR-based methods. On the other larger sequences, NICE-SLAM failed to track the scene. This reflects that NICE-SLAM was developed for small indoor scenarios and does not scale to more challenging scenes.

LONER with the CLONeR loss achieves trajectory accuracy similar to \algname on some sequences. However, it consistently performs worse, especially in Quad. 
We found that the URF loss degrades the performance of LONER
%LONER using the URF loss is less competitive, indicating that the L2 LOS loss is not suitable for real-time SLAM. This is likely because the L2 norm heavily prioritizes outliers early in training, preventing the gradient from focusing on much of the geometry reconstruction. 
We also tested LONER with the KL loss proposed by DS-NeRF \cite{dsnerf}. However, it crashes when the initial pose estimation is poor because using fixed uncertainty for the goal distribution can cause numerical instability in the SLAM context.

\subsubsection{Reconstruction Evaluation}
\input{tables/map_comparison_v5}

To evaluate maps, point clouds are created by first generating a mesh, then sampling a point cloud from the mesh. To bound the size of the generated point clouds, all maps (estimated and groundtruth) are downsampled to a voxel grid size of 5cm, except for the small MCR sequence, which uses a voxel grid size of 1cm. Finally, because groundtruth maps may extend beyond the field-of-view of the sensor used to collect each sequence, we crop each groundtruth map to the geometry observed by the sensor during data collection.

Map metrics include accuracy (mean distance from each point in the estimated map to each point in the groundtruth map) and completion (mean distance from each point in the groundtruth map to each point in the estimated map) \cite{imap, niceslam}. 
Additionally, precision and recall are computed with a 0.1m threshold.
Table~\ref{tab:map-comparison} shows quantitative evaluation for map reconstruction performance. \algname performs competitively with or better than the baselines in all tests. \algname and SHINE Mapping out-perform the other baselines. Qualitatively, Fig.~\ref{fig:reconstruction} shows that SHINE and \algname estimate the most accurate maps. SHINE estimates more complete geometry, while \algname recovers finer detail and produces maps with fewer artifacts.

\begin{figure}
    \centering
    \includegraphics[width=0.95\columnwidth]{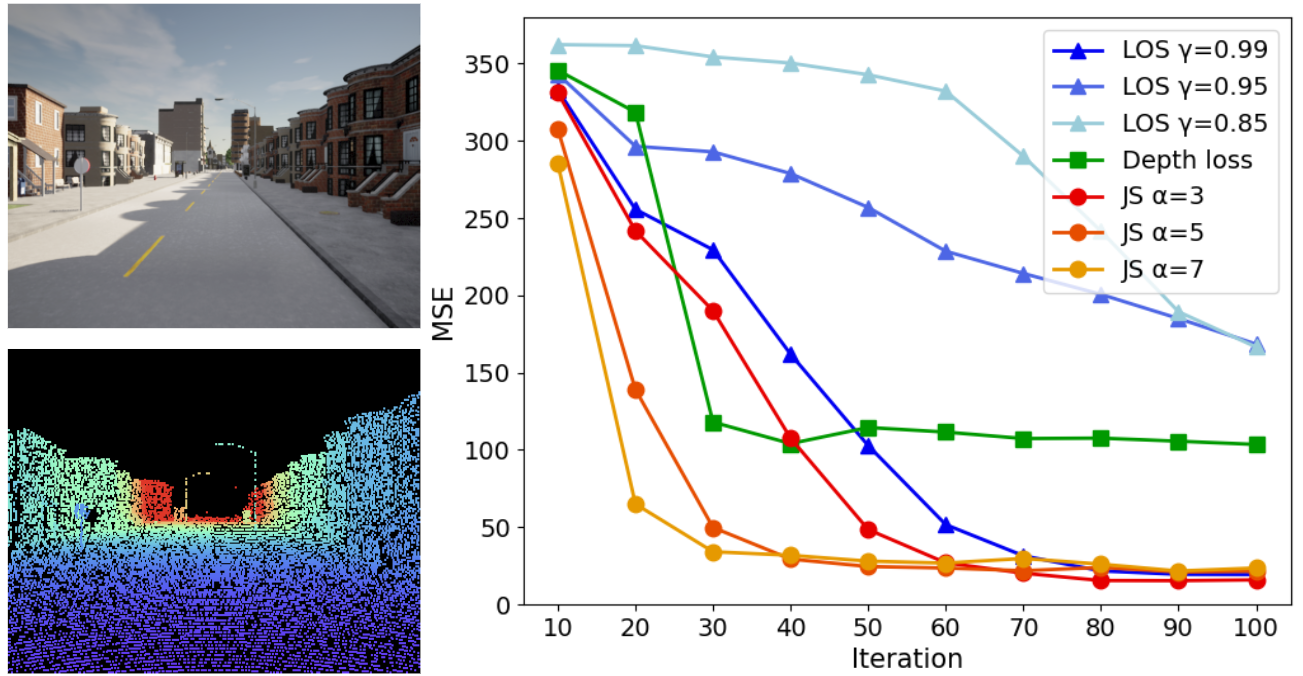}
    \caption{Training on a single LiDAR scan from the CARLA simulator indicates that the JS loss function converges faster than alternatives. The left images show simulated camera and LiDAR data. The plot on the right compares MSE (m$^2$) between groundtruth depth and estimated depth throughout training.}
    %Comparing JS loss convergence speed with other loss functions in single LiDAR scan training scenes. The LiDAR scans are collected from the CARLA simulator. The images on the right are RGB and LiDAR data from the camera perspective. The MSE is the mean square error between the camera-perspective render depth and groundtruth depth in meters.}
    \label{fig:loss_convergence}
\end{figure}

\begin{figure*}[h!t!]
    \centering
    \includegraphics[width=0.9\linewidth]{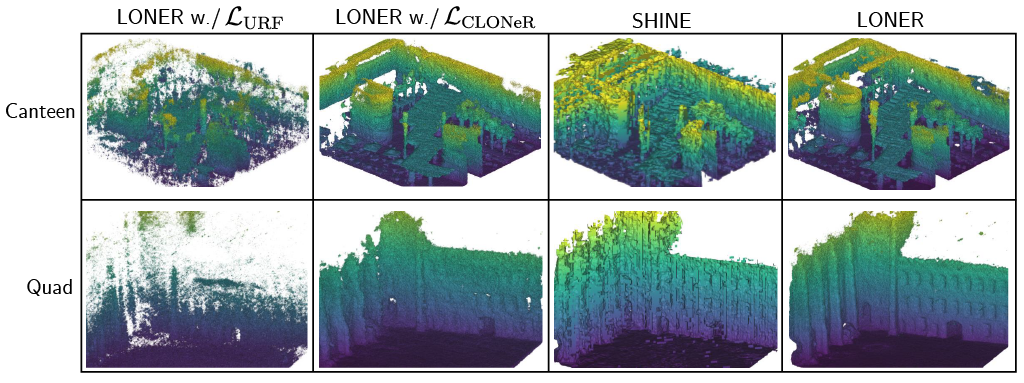}
    \caption{Reconstruction of meshes on each sequence with the benchmarked algorithms. LONER and SHINE offer the most complete and detailed results. SHINE has slightly more complete geometry, noticeable in the top-left of the Quad images where LONER omits pillars captured by SHINE. However, LONER captures details better and has fewer artifacts.}
    \label{fig:reconstruction}
\end{figure*}

\subsection{Runtime}\label{results-runtime}

Runtime performance was evaluated on a computer with an AMD Ryzen 5950X CPU and an NVidia A6000 GPU, which is similar to the platform used to benchmark NICE-SLAM~\cite{niceslam}. Each tracking step takes an average of 14ms to compute, which is faster than is needed by the 5Hz configuration. The map is updated continuously throughout operation, with 50 iterations allocated per KeyFrame and one KeyFrame added every 3 seconds. When run in parallel with the tracker, the average time to perform these 50 iterations is 2.79 seconds, or approximately 56ms per iteration. Hence, the map is updated at approximately 18Hz, and the system finishes processing a KeyFrame in under the 3 seconds allotted per KeyFrame. This ensures the system can keep up with input sensor data.

\subsection{Loss Function Performance}\label{results-loss}

\begin{figure*}
    \centering
    \includegraphics[width=0.9\textwidth]{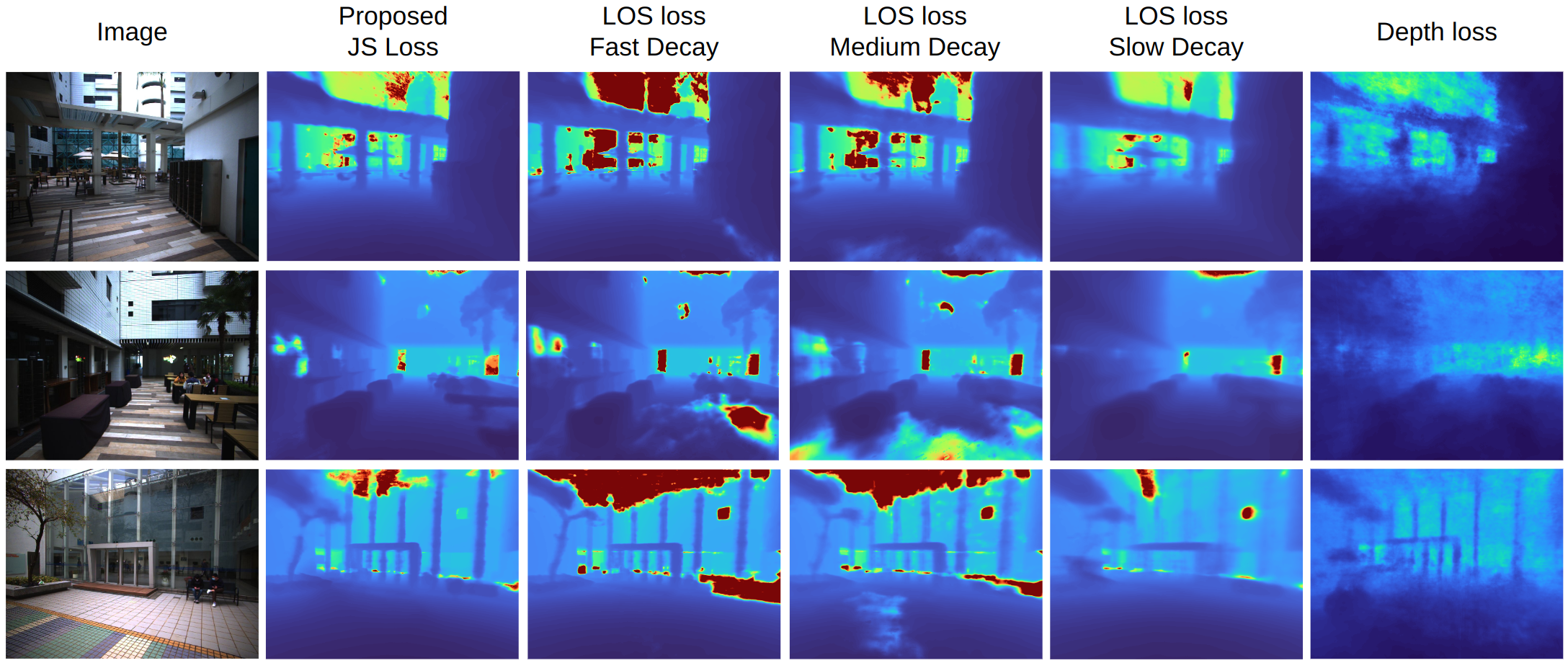}
    \caption{The depth images rendered from the MLP trained by LONER with different loss functions. The depth loss provides blurry geometry with limited training samples. The LOS loss with a fast decay rate provides more detailed geometry but worse hole-filling. In contrast, the LOS loss with a slow decay rate estimates the untrained region better but results in blurry geometry. The proposed JS loss combines the advantages of both fast and slow decay rates, which provides good hole-filling results while preserving geometry details.}
    \label{fig:depth_image_loss_comparison}
\end{figure*}

We evaluate each component of the loss function in isolation. In addition to the JS dynamic margin loss, we evaluate an LOS loss with three different exponential decay rates: 0.99 (Slow), 0.95 (Medium), and 0.85 (Fast). Finally, we consider the depth loss in isolation, as is used in \cite{niceslam, imap}. As a qualitative comparison of mapping quality, depth images rendered from each configuration are shown in Fig.~\ref{fig:depth_image_loss_comparison}. The proposed JS loss shows the most complete and detailed reconstruction of the tested configurations. %Fast LOS loss leads to sharp geometry but worse hole-filling. Slow LOS loss performs better for map completion, but it causes blurry reconstruction. By using the dynamic margin, the proposed JS loss reconstructs a crisp map while filling holes well.

Additionally, Fig.~\ref{fig:loss_convergence} demonstrates that JS loss converges faster than other losses. In this experiment, we evaluate the convergence of each function when training on a single scan in simulated data. We use the CARLA simulator, where we obtain groundtruth depth images for evaluations. We compute the mean squared error of rendered depth images throughout training to show convergence performance. The results show that our JS loss converges faster than other losses.

\subsection{Ablation Study}
This section describes the ablation studies over key components of the SLAM framework and the loss function. To compare maps in the ablations, we evaluate the L1 depth loss by comparing rendered depth to depth measured by the LiDAR. This is analogous to the L1 Depth metric commonly used in NeRF frameworks \cite{imap, niceslam, nerf-slam}. We compute the L1 depth across 25 randomly selected scans and report the mean value.

\subsubsection{SLAM Framework}

\input{tables/slam_ablation}

In Table~\ref{tab:slam_ablation}, we compare the impact of three changes to the SLAM framework. Disabling pose optimization is confirmed to strongly impact localization and mapping. Replacing the random KeyFrame selection with either the $N_W$ most recent or $N_W/2$ recent KeyFrames and $N_W/2$ randomly selected KeyFrames generally reduces performance. Finally, on the outdoor dataset, disabling sky segmentation has little effect on localization but degrades reconstruction accuracy.
% First, pose optimization is disabled, meaning poses are estimated by ICP only. Results confirm that joint optimization drastically improves trajectory estimation and reconstruction accuracy. Next, we disable sky segmentation for Quad, which is the only sequence with view of the sky. This shows similar trajectory accuracy, but significantly degraded mapping accuracy. Finally, we consider modifying the KeyFrame selection strategy to either use the most recent $N_W$ KeyFrames, or using $N_W/2$ recent KeyFrames and $N_W/2$ randomly selected KeyFrames. The fully random strategy results in the most consistent performance across all sequences.

\subsubsection{Loss Function}\label{results-loss-ablation}
\input{tables/loss_comparison_v5}

Finally, we consider disabling features of the proposed loss function, which includes both the JS loss and the depth loss. In Table~\ref{tab:loss-comparison}, we evaluate using only depth loss, depth loss, and LOS loss with the fixed medium decay rate, LOS loss with dynamic margin and no depth loss, and the full system. The results demonstrate that the proposed system performs best in all metrics on all datasets. 

%We further evaluate the impact of losses on mapping metrics. For this test, we evaluate the L1 depth loss by considering 25 randomly selected LiDAR scans from the sequence (including those withheld from training) and comparing the estimated depth to the groundtruth depth. For each trial, the overall L1 depth is calculated as the mean across the 25 trials. In this test, on all lite mode sequences and most full mode sequences, the JS loss outperforms all baselines. 

\section{Conclusions and future work} \label{conclusion}

This paper proposed \algname, the first real-time LiDAR SLAM algorithm with an implicit neural map representation. To achieve SLAM in real-time, we presented a novel loss function for depth-supervised training. Results demonstrated that the JS loss outperforms current loss functions in both reconstruction accuracy and hole-filling while maintaining low computational costs. By testing this method on public datasets, we demonstrated that \algname achieves state-of-the-art map and trajectory quality, while providing an implicit geometry representation to support novel view depth rendering.

There are several avenues of future work to continue improving \algname. First, adding RGB data without compromising runtime performance would aid in the realism of reconstructions. Additionally, considering alternate input feature embeddings and ray selection heuristics could improve the ability of \algname to operate in city-scale scenarios. Further, inertial data could help the system track accurately under rapid rotation and in feature-sparse scenarios, where the LiDAR data is less informative. Finally, to function in highly dynamic environments, more work is needed to handle dynamic objects in the scene.

%% Use plainnat to work nicely with natbib. 
%\FloatBarrier
\small
\bibliographystyle{IEEEtran}
\bibliography{references}

\end{document}

%% file: tables/configurations.tex
\begin{table}[t]
\begin{center}
\caption{Parameters for \algname.}
 \label{tab:configurations} 
\begin{tabular}{l @{\hskip 1cm}c @{\hskip 1cm}c}
        \hline
        Description& Symbol & Value \\
        \hline
        Time per KeyFrame & $t_{KF}$ & 3 \\
        KeyFrame Window Size & $N_W$ & 8\\
        Rays per KeyFrame & $N_R$ & 512 \\
        Samples per Ray & $N_S$ & 512 \\
        Min Depth Margin & $\epsilon_{min}$ & 0.5 \\
        JS Scale Hyperparameter& $\alpha$ & 1 \\
        Min/Max JS Divergence & $JS_{min}, JS_{max}$ & 1, 10 \\
        Loss Coefficients & $\lambda_{1}, \lambda_2$ & $5\times 10^{-6}$, 1 \\ 
        \hline
\end{tabular}
\end{center}
\end{table}

%% file: tables/tracking_comparison_compact.tex
\begin{table}  
\begin{center}
\setlength\tabcolsep{3.5pt}
\caption{Pose tracking results on Fusion Portable and Newer College sequences. Reported metric is RMS APE (m). An \fail indicates the algorithm failed.}
\label{tab:tracking-comparison-compact}
\begin{tabular}{l | c c c c}
& \textbf{MCR} & \textbf{Canteen} & \textbf{Garden} & \textbf{Quad} \\
\hline
\textbf{LeGO-LOAM} & 0.052&	0.129&	0.161&	\textbf{0.126}\\
\textbf{NICE-SLAM} & 0.248&	\fail	   & \fail	   & -     \\
\hline
\textbf{LONER w./ $\mathcal{L}_{\text{URF}}$}      & 0.047&	0.952&	0.928&	0.931\\
\textbf{LONER w./ $\mathcal{L}_{\text{CLONeR}}$}   & 0.034&	0.071&	0.073&	0.306\\
\hline
\textbf{\algname} & \textbf{0.029}&	\textbf{0.064}&	\textbf{0.056}&	0.130\\
\hline
\end{tabular}
\end{center}
\end{table}

%% file: tables/map_comparison_v5.tex
\begin{table}
  \centering
  \caption{Comparison of map Acccuracy (m), Completion (m), Precision, and Recall between proposed and baseline algorithms. Unlike the others, SHINE used ground truth poses. A `\na' indicates invalid configurations, while `\fail' indicates that the algorithm failed.}
  \label{tab:map-comparison}
  \begin{tabular}{l@{\hskip2pt} l | c c | c c | c}

  % && \multicolumn{2}{c|}{\textbf{Baselines}} & \multicolumn{2}{c|}{\textbf{Basline Loss}} & \multirow{3}{*}{\textbf{\algname}}\\
  % \cline{3-6}
  && \textbf{NICE} &  \multirow{2}{*}{\textbf{SHINE}}  & \scriptsize{\textbf{LONER w./}}&\scriptsize\textbf{LONER w./} & \multirow{2}{*}{\textbf{\algname}}\\
  &&\textbf{SLAM} & & \textbf{$\mathbf{\mathcal{L}}_{\text{CLONeR}}$}& \textbf{$\mathbf{\mathcal{L}}_{\text{URF}}$}&  \\
    \hline
    \multirow{4}{*}{\verttab{\textbf{MCR}}}
    &Acc. & 0.621 &0.164  &  \textbf{0.110}  & 0.153   & 0.186 \\
    &Cmp.  & 0.419 &0.075  & 0.080   & 0.102  & \textbf{0.069} \\
    &Prec.   & 0.124 &0.624  & \textbf{0.665}	& 0.449   & 0.473 \\
    &Rec.    & 0.476  &0.757  & \textbf{0.940}& 0.884  & 0.932 \\
    \hline
    \multirow{4}{*}{\verttab{\textbf{Canteen}}}
    &Acc.  &\multirow{4}{*}{\fail} & \na & \na & \na & \na \\
    &Cmp. & &0.116 & 0.220	& 0.190  & \textbf{0.105} \\
    &Prec.  & & \na & \na & \na & \na \\
    &Rec.   &  &0.753 & 0.524	& 0.846  & \textbf{0.878} \\
    \hline
    \multirow{4}{*}{\verttab{\textbf{Garden}}}
    &Acc. & \multirow{4}{*}{\fail} & \na & \na & \na & \na \\
    &Cmp. & &\textbf{0.130} & 0.333	& 0.539  & 0.157 \\
    &Prec.  & & \na & \na & \na & \na \\
    &Rec.   &  &0.657 & 0.469	& 0.623 & \textbf{0.784} \\
    \hline
    \multirow{4}{*}{\verttab{\textbf{Quad}}}
    &Acc.  & \na & \textbf{0.301} & 0.663	& 0.552 & 0.380 \\
    &Cmp. & \na & \textbf{0.148} & 0.543	& 0.895 & 0.373 \\
    &Prec.  & \na  & \textbf{0.453} & 0.150	& 0.127 & 0.327 \\
    &Rec.   & \na & 0.717 & 0.602	& 0.484 & \textbf{0.809} \\
  \end{tabular}
\end{table}

%% file: tables/slam_ablation.tex
\begin{table*}[h!t!]
  \centering
  \caption{We perform an ablation study over the SLAM framework by disabling key features, and show the proposed system outperforms alternatives.}
  \label{tab:slam_ablation}
  \begin{tabular}{c c c| c c | c c | c c | c c}
  \textbf{Pose} & \textbf{Sky} & \textbf{Random}& \multicolumn{2}{c|}{\textbf{MCR}} & \multicolumn{2}{c|}{\textbf{Canteen}}
   &\multicolumn{2}{c|}{\textbf{Garden}} & \multicolumn{2}{c}{\textbf{Quad}}\\
  \textbf{Optimization} & \textbf{Segmentation}  & \textbf{KeyFrames}& $t_{APE}$ & L1 Depth & $t_{APE}$ & L1 Depth & $t_{APE}$ & L1 Depth & $t_{APE}$ & L1 Depth\\
  \hline
\emptycirc& \fullcirc& \fullcirc & 0.077&	0.317&	1.162&	2.927&	1.223&	3.305&	0.862&	3.302 \\
\fullcirc& \emptycirc& \fullcirc & - &	- &	- &	- & - &	- &	\textbf{0.128}&	0.994 \\
\fullcirc& \fullcirc& \emptycirc & \textbf{0.028}&	0.289&	0.079&	1.573&	0.066&	1.237&	0.219&	1.363 \\
\fullcirc& \fullcirc& \halfcirc  & 0.029&	0.285&	0.065&	1.298&	0.057&	1.261& 0.130&	\textbf{0.829} \\ 
\hline
\fullcirc& \fullcirc& \fullcirc  & 0.029&	\textbf{0.284}&	\textbf{0.064}&	\textbf{1.296}&	\textbf{0.056}&	\textbf{1.198}&	0.130 &	0.880
  \end{tabular}
\end{table*}

%% file: tables/loss_comparison_v5.tex
% \begin{table}
%   \centering
%   \caption{We perform an ablation study over the loss by disabling the dynamic margin and the depth loss individually.}
%   \label{tab:loss-comparison}
%   \begin{tabular}{c@{\hskip5pt}c | l | c@{\hskip5pt}c@{\hskip5pt}c@{\hskip5pt}c}
%   \textbf{Depth} & \textbf{Dynamic} & \multirow{2}{*}{\textbf{Metric}} & \multirow{2}{*}{\textbf{MCR}} & \multirow{2}{*}{\textbf{Canteen}} & \multirow{2}{*}{\textbf{Garden}} & \multirow{2}{*}{\textbf{Quad}}\\
%   \textbf{Loss} & \textbf{Margin}  & & & & & \\
%   \hline
% \multirow{2}{*}{\fullcirc}& \multirow{2}{*}{\emptycirc} &   ${t_{APE}}$ & 0.033 & 0.075 & 0.076 & 0.570\\
%      &&                                             {L1 Depth}    & 0.338 & 1.453 & 1.304 & 1.747\\
%      \hline
% \multirow{2}{*}{\emptycirc}& \multirow{2}{*}{\fullcirc} &   ${t_{APE}}$   & 0.030 & 0.068 & \textbf{0.056} & 0.154\\
%                                   &&                L1 Depth    & 0.358 & 1.907 & 1.490 & 2.262\\
%     \Xhline{3\arrayrulewidth}
%   \multirow{2}{*}{\fullcirc}& \multirow{2}{*}{\fullcirc} &${t_{APE}}$  & \textbf{0.029} & \textbf{0.064} & \textbf{0.056} & \textbf{0.130}\\
%                                                &  & {L1 Depth}    & \textbf{0.284} & \textbf{1.296} & \textbf{1.198} & \textbf{0.880}\\
%   \end{tabular}
% \end{table}

\begin{table*}[h!t!]
  \centering
  \caption{We perform an ablation study over the loss. The first row uses only depth loss. The second uses depth loss and LOS loss with no dynamic margin. The third row uses LOS Loss with dynamic margin. The final row is the proposed system.}
  \label{tab:loss-comparison}
  \begin{tabular}{c c c | c c | c c | c c | c c}
  \textbf{Depth} & \textbf{LOS} &\textbf{Dynamic} & \multicolumn{2}{c|}{\textbf{MCR}} & \multicolumn{2}{c|}{\textbf{Canteen}}
   &\multicolumn{2}{c|}{\textbf{Garden}} & \multicolumn{2}{c}{\textbf{Quad}}\\
  \textbf{Loss} & \textbf{Loss} & \textbf{Margin} & $t_{APE}$ & L1 Depth & $t_{APE}$ & L1 Depth & $t_{APE}$ & L1 Depth & $t_{APE}$ & L1 Depth\\
  \hline
\fullcirc& \emptycirc& N/A          & 0.046&	0.355&	1.236&	3.014&	0.788&	2.447&	0.779&	2.265 \\
\fullcirc& \fullcirc& \emptycirc    & 0.033&	0.338&	0.075&	1.453&	0.076&	1.304&	0.570&	1.747 \\
\emptycirc & \fullcirc & \fullcirc  & 0.030&	0.358&	0.068&	1.907&	\textbf{0.056}&	1.490&	0.154&	2.262 \\
\hline
\fullcirc& \fullcirc& \fullcirc  & \textbf{0.029}&	\textbf{0.284}&	\textbf{0.064}&	\textbf{1.296}&	\textbf{0.056}&	\textbf{1.198}&	\textbf{0.130}&	\textbf{0.880}
  \end{tabular}
\end{table*}

%% file: loner.bbl
\begin{thebibliography}{10}
\providecommand{\url}[1]{#1}
\csname url@rmstyle\endcsname
\providecommand{\newblock}{\relax}
\providecommand{\bibinfo}[2]{#2}
\providecommand\BIBentrySTDinterwordspacing{\spaceskip=0pt\relax}
\providecommand\BIBentryALTinterwordstretchfactor{4}
\providecommand\BIBentryALTinterwordspacing{\spaceskip=\fontdimen2\font plus
\BIBentryALTinterwordstretchfactor\fontdimen3\font minus
  \fontdimen4\font\relax}
\providecommand\BIBforeignlanguage[2]{{%
\expandafter\ifx\csname l@#1\endcsname\relax
\typeout{** WARNING: IEEEtran.bst: No hyphenation pattern has been}%
\typeout{** loaded for the language `#1'. Using the pattern for}%
\typeout{** the default language instead.}%
\else
\language=\csname l@#1\endcsname
\fi
#2}}

\bibitem{nerf}
B.~Mildenhall, P.~P. Srinivasan, M.~Tancik, J.~T. Barron, R.~Ramamoorthi, and
  R.~Ng, ``Nerf: Representing scenes as neural radiance fields for view
  synthesis,'' \emph{Commun. ACM}, vol.~65, no.~1, p. 99–106, Dec 2021.

\bibitem{nerf-navigation}
M.~Adamkiewicz, \emph{et~al.}, ``Vision-{Only} {Robot} {Navigation} in a
  {Neural} {Radiance} {World},'' \emph{IEEE Robotics and Automation Letters},
  vol.~7, no.~2, pp. 4606--4613, Apr. 2022.

\bibitem{active-nerf}
X.~Pan, Z.~Lai, S.~Song, and G.~Huang, ``Activenerf: Learning where to see with
  uncertainty estimation,'' in \emph{2022 European Conference on Computer
  Vision}.\hskip 1em plus 0.5em minus 0.4em\relax Springer, 2022, pp. 230--246.

\bibitem{fusion-portable}
J.~Jiao, \emph{et~al.}, ``Fusionportable: A multi-sensor campus-scene dataset
  for evaluation of localization and mapping accuracy on diverse platforms,''
  \emph{IEEE/RSJ International Conference on Intelligent Robots and Systems},
  pp. 3851--3856, 2022.

\bibitem{imap}
E.~Sucar, S.~Liu, J.~Ortiz, and A.~J. Davison, ``imap: Implicit mapping and
  positioning in real-time,'' \emph{2021 IEEE/CVF International Conference on
  Computer Vision}, pp. 6209--6218, 2021.

\bibitem{niceslam}
Z.~Zhu, \emph{et~al.}, ``Nice-slam: Neural implicit scalable encoding for
  slam,'' in \emph{IEEE/CVF Conference on Computer Vision and Pattern
  Recognition}, 2022.

\bibitem{nerf-slam}
A.~Rosinol, J.~J. Leonard, and L.~Carlone, ``Nerf-slam: Real-time dense
  monocular slam with neural radiance fields,'' \emph{ArXiv}, vol.
  abs/2210.13641, 2022.

\bibitem{urf}
K.~Rematas, \emph{et~al.}, ``Urban radiance fields,'' \emph{IEEE/CVF Conference
  on Computer Vision and Pattern Recognition}, 2022.

\bibitem{cloner}
A.~Carlson, M.~S. Ramanagopal, N.~Tseng, M.~Johnson-Roberson, R.~Vasudevan, and
  K.~A. Skinner, ``Cloner: Camera-lidar fusion for occupancy grid-aided neural
  representations,'' \emph{IEEE Robotics and Automation Letters}, vol.~8,
  no.~5, pp. 2812--2819, 2023.

\bibitem{shinemapping}
X.~Zhong, Y.~Pan, J.~Behley, and C.~Stachniss, ``Shine-mapping: Large-scale 3d
  mapping using sparse hierarchical implicit neural representations,''
  \emph{arXiv preprint arXiv:2210.02299}, 2022.

\bibitem{nerfloam}
J.~Deng, \emph{et~al.}, ``Nerf-loam: Neural implicit representation for
  large-scale incremental lidar odometry and mapping,'' \emph{arXiv preprint
  arXiv:2303.10709}, 2023.

\bibitem{yan2023}
D.~Yan, X.~Lyu, J.~Shi, and Y.~Lin, ``Efficient implicit neural reconstruction
  using lidar,'' \emph{arXiv preprint arXiv:2302.14363}, 2023.

\bibitem{loam}
J.~Zhang and S.~Singh, ``Loam: Lidar odometry and mapping in real-time,'' in
  \emph{Robotics: Science and Systems}, 2014.

\bibitem{suma}
J.~Behley and C.~Stachniss, ``Efficient surfel-based slam using 3d laser range
  data in urban environments.'' in \emph{Robotics: Science and Systems}, 2018.

\bibitem{lego-loam}
T.~Shan and B.~Englot, ``Lego-loam: Lightweight and ground-optimized lidar
  odometry and mapping on variable terrain,'' in \emph{IEEE/RSJ International
  Conference on Intelligent Robots and Systems}, 2018, pp. 4758--4765.

\bibitem{lio-mapping}
H.~Ye, Y.~Chen, and M.~Liu, ``Tightly coupled 3d lidar inertial odometry and
  mapping,'' in \emph{IEEE International Conference on Robotics and
  Automation}, 2019, pp. 3144--3150.

\bibitem{lio-sam}
T.~Shan, B.~Englot, D.~Meyers, W.~Wang, C.~Ratti, and D.~Rus, ``Lio-sam:
  Tightly-coupled lidar inertial odometry via smoothing and mapping,'' in
  \emph{IEEE/RSJ International Conference on Intelligent Robots and Systems},
  2020, pp. 5135--5142.

\bibitem{droid-slam}
Z.~Teed and J.~Deng, ``\BIBforeignlanguage{English (US)}{Droid-slam: Deep
  visual slam for monocular, stereo, and rgb-d cameras},'' in
  \emph{\BIBforeignlanguage{English (US)}{34 - 35th Conference on Neural
  Information Processing Systems}}.\hskip 1em plus 0.5em minus 0.4em\relax
  Neural information processing systems foundation, 2021, pp. 16\,558--16\,569.

\bibitem{meslam}
E.~Kruzhkov, \emph{et~al.}, ``Meslam: Memory efficient slam based on neural
  fields,'' in \emph{2022 IEEE International Conference on Systems, Man, and
  Cybernetics}, 2022, pp. 430--435.

\bibitem{co-slam}
H.~Wang, J.~Wang, and L.~Agapito, ``Co-slam: Joint coordinate and sparse
  parametric encodings for neural real-time slam,'' in \emph{IEEE/CVF
  Conference on Computer Vision and Pattern Recognition}, 2023, pp.
  13\,293--13\,302.

\bibitem{eslam}
M.~M. Johari, C.~Carta, and F.~Fleuret, ``Eslam: Efficient dense slam system
  based on hybrid representation of signed distance fields,'' in \emph{IEEE/CVF
  Conference on Computer Vision and Pattern Recognition}, 2023, pp.
  17\,408--17\,419.

\bibitem{vox-fusion}
X.~Yang, H.~Li, H.~Zhai, Y.~Ming, Y.~Liu, and G.~Zhang, ``Vox-fusion: Dense
  tracking and mapping with voxel-based neural implicit representation,'' in
  \emph{2022 IEEE International Symposium on Mixed and Augmented Reality},
  2022, pp. 499--507.

\bibitem{dsnerf}
K.~Deng, A.~Liu, J.-Y. Zhu, and D.~Ramanan, ``Depth-supervised nerf: Fewer
  views and faster training for free,'' in \emph{IEEE/CVF Conference on
  Computer Vision and Pattern Recognition}, 2022, pp. 12\,882--12\,891.

\bibitem{orbslam}
R.~Mur-Artal, J.~M.~M. Montiel, and J.~D. Tardós, ``Orb-slam: A versatile and
  accurate monocular slam system,'' \emph{IEEE Transactions on Robotics},
  vol.~31, no.~5, pp. 1147--1163, 2015.

\bibitem{icp}
S.~Rusinkiewicz and M.~Levoy, ``Efficient variants of the icp algorithm,'' in
  \emph{Proceedings Third International Conference on 3-D Digital Imaging and
  Modeling}, 2001, pp. 145--152.

\bibitem{instant-ngp}
T.~M\"uller, A.~Evans, C.~Schied, and A.~Keller, ``Instant neural graphics
  primitives with a multiresolution hash encoding,'' \emph{ACM Trans. Graph.},
  vol.~41, no.~4, pp. 102:1--102:15, July 2022.

\bibitem{JS}
J.~Lin, ``Divergence measures based on the shannon entropy,'' \emph{IEEE
  Transactions on Information Theory}, vol.~37, no.~1, pp. 145--151, 1991.

\bibitem{newer-college}
M.~Ramezani, Y.~Wang, M.~Camurri, D.~Wisth, M.~Mattamala, and M.~Fallon, ``The
  newer college dataset: Handheld lidar, inertial and vision with ground
  truth,'' in \emph{IEEE/RSJ International Conference on Intelligent Robots and
  Systems}, 2020, pp. 4353--4360.

\bibitem{raft}
Z.~Teed and J.~Deng, ``Raft: Recurrent all-pairs field transforms for optical
  flow,'' in \emph{European Conference on Computer Vision}, A.~Vedaldi,
  H.~Bischof, T.~Brox, and J.-M. Frahm, Eds.\hskip 1em plus 0.5em minus
  0.4em\relax Springer, 2020, pp. 402--419.

\bibitem{traj-evaluation}
Z.~Zhang and D.~Scaramuzza, ``A tutorial on quantitative trajectory evaluation
  for visual(-inertial) odometry,'' in \emph{IEEE/RSJ International Conference
  on Intelligent Robots and Systems}, 2018, pp. 7244--7251.

\end{thebibliography}
